\documentclass[letterpaper, 10 pt, conference]{ieeeconf}

\IEEEoverridecommandlockouts

\usepackage{amsmath}
\usepackage{amssymb}
\usepackage{bm}
\usepackage{bbm}
\usepackage{booktabs}
\usepackage{tabularx}
\usepackage{graphicx}
\usepackage{algorithm}
\usepackage{algpseudocode}
\usepackage{subcaption}
\usepackage{cite}
\usepackage{multirow}
\makeatletter
\let\NAT@parse\undefined
\makeatother
\usepackage{hyperref}

\newcommand{\sg}{\operatorname{sg}}
\newcommand{\mb}[1]{\mathbf{#1}}
\newcommand{\mr}[1]{\mathrm{#1}}
\newcommand{\mc}[1]{\mathcal{#1}}

\title{\LARGE \bf
CQDC-Warp: GPU-Accelerated Differentiable Quasi-Dynamic Simulation for Contact-Rich Manipulation
}

\author{\authorblockN{Yanchen Ding$^{1,2*}$,
Jiayi Chen$^{1,2*}$,
Jianfeng Xu$^{1,2}$, 
Wei Zhao$^{2,3}$,
Li Yi$^{2,3\dagger}$,
He Wang$^{1,2\dagger}$}
\authorblockA{$^{1}$Peking University, $^2$Galbot, $^3$Tsinghua University}
\authorblockA{$^*$Equal contribution, $^\dagger$Corresponding author}
}

\begin{document}
\bstctlcite{cqdc:reference-style}

\maketitle
\thispagestyle{empty}
\pagestyle{empty}

\suppressfloats[t]

\vspace{-1em}
\begin{abstract}
Efficient synthesis of contact-rich manipulation trajectories is essential for
downstream robot learning. Gradient-based planning offers a promising approach,
but jointly optimizing motion and contact decisions remains challenging.
Meanwhile, many existing planners run on CPUs with limited throughput.
We present \emph{CQDC-Warp}, a GPU-accelerated differentiable quasi-dynamic
simulator for batched model predictive control (MPC). We use it to study
contact-guided trajectory optimization, in which intended robot--object
contacts are specified before planning and encouraged through soft costs,
while the simulated dynamics
determine contact geometry and frictional response without a prescribed
contact-mode schedule. Across
$17{,}408$ trials on the $17$-object Allegro in-hand rotation benchmark, our
framework achieves $79.8\%$ success, compared with $73.7\%$ for CFree and
$65.2\%$ for IMPACT, and reduces mean trajectory-generation time by
factors of $27.0$ and $50.2$, respectively. On a single NVIDIA RTX 4090 with
$2{,}048$ parallel worlds, forward and backward throughput reach
$2.06\times10^5$ and $2.25\times10^5$ world-steps per second, respectively.
Further experiments cover manipulation with LEAP and Shadow hands,
environmental contact exploitation, bimanual manipulation, whole-body control,
and real-world trajectory replay on a Sharpa hand.
\end{abstract}

\begin{figure}[t]
 \centering
 \includegraphics[width=\columnwidth]{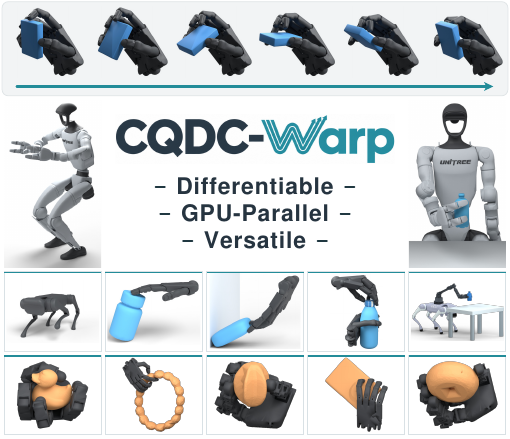}
 \caption{\textbf{CQDC-Warp.} A differentiable contact simulator for
 GPU-parallel trajectory generation on versatile robotic tasks.}
 \label{fig:teaser}
 \vspace{-5mm}
\end{figure}

\section{Introduction}
\label{sec:introduction}

\begin{table*}[!t]
\centering
\footnotesize
\setlength{\tabcolsep}{2.5pt}
\renewcommand{\arraystretch}{1.22}
\begin{tabular*}{0.9\textwidth}{@{\extracolsep{\fill}}lcccccccc@{}}
\toprule
Simulator & Year & Device & Dynamics & \begin{tabular}{@{}c@{}}Contact\\Model\end{tabular}
& \begin{tabular}{@{}c@{}}Dynamics\\Gradient\end{tabular}
& Geometry & \begin{tabular}{@{}c@{}}Geometry\\Gradient\end{tabular}
& \begin{tabular}{@{}c@{}}Demonstrated\\Robots\end{tabular} \\
\midrule
ADD~\cite{geilinger2020add}
& 2020 & CPU & Full & Penalty & Implicit & SDF & Analytic
& UR5 \\
Nimble~\cite{werling2021fast}
& 2021 & CPU & Full & LCP & Implicit & Meshes & Analytic
& Catapult \\
Dojo~\cite{howell2022dojo}
& 2022 & CPU & Full & NCP & Implicit & Primitives & Analytic
& Ant / HalfCheetah \\
Silico~\cite{lecleach2023singlelevel}
& 2023 & CPU & Full & NCP & Implicit & Primitives & Implicit
& Fingertips \\
CQDC~\cite{pang2022global}
& 2023 & CPU & Quasi-dynamic & CCP & Implicit & Primitives & AutoDiff
& Allegro / Dual iiwa \\
Simple~\cite{lelidec2025simple,carpentier2024simple}
& 2025 & CPU & Full & NCP & Implicit & Meshes & Surrogate
& CartPole / Hopper \\
ComFree~\cite{jin2024complementarity}
& 2025 & CPU & Quasi-dynamic & Penalty & AutoDiff & Meshes & -
& TriFinger / Allegro \\
\midrule
TDS~\cite{heiden2021neuralsim,tdssoftware}
& 2021 & GPU & Full & NCP & AutoDiff & Primitives & AutoDiff
& - \\
DiffSDFSim~\cite{strecke2021diffsdfsim}
& 2021 & GPU & Full & LCP & Implicit & SDF & AutoDiff
& - \\
Brax~\cite{freeman2021brax}
& 2021 & GPU & Full & Impulse & AutoDiff & Primitives & AutoDiff
& Reacher \\
DiffMJX~\cite{paulus2026differentiable}
& 2026 & GPU & Full & CCP & AutoDiff & Meshes & AutoDiff
& MyoHand / MyoArm \\
\textbf{CQDC-Warp (ours)}
& 2026 & GPU & Quasi-dynamic & CCP & Implicit & Meshes & Surrogate
& \begin{tabular}{@{}c@{}}Allegro / LEAP / Shadow\\Sharpa / GO2 / A1 / G1\end{tabular} \\
\bottomrule
\end{tabular*}
\caption{\textbf{Representative differentiable rigid-body contact simulators.}
CQDC-Warp combines GPU-batched simulation with surrogate geometry gradients
for meshes and demonstrates manipulation across diverse robot platforms.}
\label{tab:differentiable_simulation_comparison}
\vspace{-3mm}
\end{table*}

Contact-rich dexterous manipulation is a central capability for
general-purpose embodied intelligence.  Modern robot learning increasingly
relies on large, diverse, and high-quality datasets of robot
behavior~\cite{openx2024}.  Collecting such data for dexterous hands is
particularly challenging due to the high cost and complexity of teleoperation,
as well as differences between human and robot
embodiments~\cite{wang2024dexcap,jiang2025dexmimicgen}.  Scalable generation of
physically plausible dexterous trajectories could therefore support both
task execution and demonstration collection for downstream robot learning.

Reinforcement learning (RL) has demonstrated impressive dexterous manipulation
behaviors~\cite{rajeswaran2017learning,andrychowicz2020learning,
akkaya2019solving} covering a broad range of objects~\cite{chen2022system,
qi2023rotateit,yang2025anyrotate}.  However, acquiring these behaviors with
model-free RL typically requires extensive trial and error guided by
task-specific rewards.  The underlying dynamics are treated as a black box:
policy updates rely on sampled experience rather than derivatives of the
physical dynamics.  Moreover, learned policies remain sensitive to their
training distributions, which can limit their reuse across contact-rich
tasks and scenes.

Gradient-based planning provides another route, exploiting dynamics
derivatives to efficiently refine action
sequences~\cite{kelly2017trajectory,tassa2012synthesis}.  Within this paradigm,
contact-implicit trajectory optimization is an active and promising research
direction: it jointly optimizes state and control trajectories and contact
interactions without prescribing intended contact in advance~\cite{posa2014direct,
sleiman2021contactimplicit,kim2025contact,li2025surprising,li2026impact}.
However, this generality also introduces strong coupling between motion and
contact decisions, resulting in large, nonconvex optimization problems that
can be difficult to solve reliably.  Yet, from a practical standpoint, an even
more basic question remains open: given a suitable initialization and intended
contacts specified in advance, can existing tools reliably optimize action
sequences across diverse robot platforms and contact-rich tasks? Here, intended contacts 
specify which robot link should touch the object, rather than prescribing a complete 
contact-mode schedule; their exact geometry and frictional response are still determined 
by the simulated dynamics. This assumption is practical in many manipulation settings, 
where intended contacts can be inferred from task structure, grasp or trajectory initialization, 
or human demonstrations with tactile gloves.

Differentiable simulation supports such local optimization by providing
gradients through simulated rollouts.  Existing differentiable contact
simulators, however, face several limitations.  Some methods introduce
force-at-a-distance through smoothing~\cite{pang2022global} or neglect coupling
between contact constraints~\cite{jin2024complementarity}.  Many implementations run on CPUs
or optimize only one trajectory at a time~\cite{jin2024complementarity,lecleach2024fast,
li2026impact}, limiting throughput.  GPU-based differentiable simulators are
available~\cite{heiden2021neuralsim,freeman2021brax,paulus2026differentiable},
but their published evaluations typically cover only one or two robot
embodiments and a limited set of tasks.  Efficient trajectory generation across
diverse robot platforms and contact-rich tasks remains less explored.

In this work, we propose \emph{CQDC-Warp}, a GPU-accelerated differentiable
simulator that extends the Convex Quasi-Dynamic Contact (CQDC) model in three
respects.  First, we introduce compliant frictional contact to regularize the
contact solve and revise the inertial model to retain coupling between
actuated and unactuated coordinates.  Second, we provide implicit derivatives
of the contact dynamics and sampled surrogate derivatives of contact geometry.
Third, we develop an efficient GPU implementation built on MuJoCo-Warp's
batched pipeline~\cite{mujocowarp2025}, enabling parallel model predictive
control (MPC) across independent simulation worlds. Experiments demonstrate
in-hand manipulation, environmental contact exploitation, bimanual
manipulation, and whole-body control across dexterous hands, quadrupeds, and
humanoids, as well as a trajectory replay in the real world.

Our main contributions are:
\begin{itemize}
  \item CQDC-Warp, a GPU-accelerated differentiable quasi-dynamic contact simulator,
  integrated into a batched MPC framework.
  \item A contact-guided trajectory optimization setting in which intended 
  contacts are specified before planning, while the exact contact realization 
  remains implicit.
  \item Experimental validation across diverse robots and
  contact-rich tasks, including simulation benchmarks and
  real-world trajectory replay.
\end{itemize}

\section{Related Work}
\label{sec:related_work}

\subsection{Rigid-Body Contact Models}

Classical penalty models prescribe local contact forces that penalize
penetration, typically using spring--damper laws for the normal response
and regularized friction~\cite{geilinger2020add,heiden2021neuralsim}.
Larger penalty stiffness reduces penetration but can worsen numerical
conditioning~\cite{geilinger2020add}.
Complementarity-based formulations instead express rigid frictional contact
as a nonlinear complementarity problem (NCP), combining the Signorini
condition, Coulomb friction, and
maximum dissipation~\cite{lelidec2024contact,howell2022dojo,
lecleach2023singlelevel,lelidec2025simple,carpentier2024simple,
heiden2021neuralsim}.
Polyhedral approximations of friction cones yield linear complementarity
problems (LCPs)~\cite{stewart1996implicit,werling2021fast,strecke2021diffsdfsim}.
Cone complementarity problems (CCPs) retain the circular friction cones but
relax normal complementarity, enabling convex optimization at the cost of
possible gliding artifacts~\cite{anitescu2006optimization,lelidec2024contact};
this formulation underlies CQDC~\cite{pang2022global}.

\subsection{Differentiable Contact Simulation}

Dynamics gradients can be obtained through automatic differentiation (AD),
which applies the chain rule to the simulator's computational
graph~\cite{heiden2021neuralsim,freeman2021brax}.
For iterative contact solvers, unrolling makes differentiation cost grow
with the iteration count~\cite{lelidec2025simple}.
Implicit differentiation instead linearizes the contact equations or
optimality conditions at the converged solution and solves a sensitivity
system, avoiding differentiation through solver
iterations~\cite{werling2021fast,howell2022dojo,geilinger2020add,
pang2022global,lelidec2025simple}.
CQDC-Warp follows this implicit approach.

Geometry gradients describe how contact points, normals, and gaps change
with body poses. Signed distance fields (SDFs) provide distances and
surface normals through field values and spatial derivatives, supporting
differentiable contact with nonconvex shapes~\cite{strecke2021diffsdfsim}.
However, constructing and storing an SDF adds preprocessing and
memory overhead, and geometry edits may require rebuilding the
field~\cite{physxsdfdocumentation}.
For triangle-based collision queries, analytic derivatives
and direct AD capture variation within a triangle but not transitions
between triangles~\cite{werling2021fast,chen2025robust}.
Silico obtains implicit geometry gradients by jointly solving geometry
and dynamics, but embedding geometric variables and constraints enlarges
the coupled system~\cite{lecleach2023singlelevel}.
Sampling-based surrogate gradients aggregate information across neighboring
surface features, providing informative local gradients beyond a single
triangle~\cite{montaut2023differentiable,chen2025robust}.
CQDC-Warp adopts this surrogate approach.
Table~\ref{tab:differentiable_simulation_comparison} summarizes these
differentiation choices across simulators.

\subsection{Contact-Rich Trajectory Optimization}

Trajectory optimization commonly uses single shooting, multiple shooting,
or direct collocation~\cite{kelly2017trajectory}.
Single shooting optimizes control sequences through a complete rollout;
multiple shooting introduces intermediate states and continuity constraints
between shorter rollouts~\cite{sleiman2021contactimplicit,kim2025contact}.
Direct collocation represents trajectories with piecewise polynomials and
enforces dynamics at selected nodes~\cite{kelly2017trajectory}.
The latter two allow direct state-trajectory initialization and expose
temporal sparsity, at the cost of additional variables and constraints.
Differentiable simulation also enables gradient-based single-shooting MPC
on GPUs~\cite{paulus2026differentiable}.
CQDC-Warp follows this route, optimizing command sequences through
GPU-batched differentiable rollouts.

A separate design choice is how much contact structure is prescribed.
Crocoddyl~\cite{mastalli2020crocoddyl} optimizes motion given a predefined
contact-mode sequence and enforces the corresponding contact constraints.
Contact-implicit methods instead optimize motion and
contact interactions jointly without requiring such a
sequence~\cite{posa2014direct,sleiman2021contactimplicit,kim2025contact,
li2025surprising,li2026impact}.
Our contact-guided setting uses an intermediate level of contact prior.

\section{Preliminaries}
\label{sec:preliminaries}

\subsection{CQDC: Convex Quasi-Dynamic Contact Dynamics}
\label{subsec:classical_cqdc}

Pang et al.~\cite{pang2022global} propose a convex, differentiable
quasi-dynamic contact model, written as
\begin{equation}
 \mb{Q}\delta q+b=\mb{J}^\top\lambda,
 \quad
 \mc{K}_{\mu_i}\!\ni\lambda_i\perp
 \mb{J}_i\delta q+g_i e_n\in\!\mc{K}_{\mu_i}^*,\;\; \forall\, i,
 \label{eq:cqdc_equation_mat}
\end{equation}
where
\begin{equation}
 \begin{gathered}
 \mb{Q}=\begin{bmatrix}
 \mb{K}_{\mr{a}} & \mb{0} \\
 \mb{0} & \varepsilon\mb{M}_{\mr{u}}/h^2
 \end{bmatrix},\quad
 b=\begin{bmatrix}
 \mb{K}_{\mr{a}}(q^{\mr{a}}-u)-\tau^{\mr{a}}\\
 -\tau^{\mr{u}}
 \end{bmatrix}, \\[4pt]
 \mb{J}=\begin{bmatrix}
 \mb{J}_\mr{a} & \mb{J}_\mr{u}
 \end{bmatrix}=\operatorname{col}(\mb{J}_1,\ldots,\mb{J}_{n_c}).
 \end{gathered}
 \label{eq:cqdc_QbJ_definition}
\end{equation}
Here $q^{\mr{a}}$ and $q^{\mr{u}}$ are the actuated and unactuated
configurations, $\delta q$ is the corresponding tangent displacement, $u$ is the
actuator target, and $\tau^{\mr{a}},\tau^{\mr{u}}$ are generalized
external forces.  Matrices $\mb{K}_{\mr{a}}$ and $\mb{M}_{\mr{u}}$ are the
actuator stiffness and unactuated mass matrices; $h$ is the step size, and
$\varepsilon$ controls the inertial regularization.  The stiffness block
treats the actuator as a position-controlled impedance with equilibrium
command $u$~\cite{hogan1985impedance,pang2022global}.  The
Jacobian $\mb{J}$ stacks $n_c$ per-contact Jacobians $\mb{J}_i$, with
$\mb{J}_{\mr{a}}$ and $\mb{J}_{\mr{u}}$ its actuated and unactuated
blocks, while $\lambda$ stacks the corresponding contact forces.  At
contact $i$, $g_i$ is the signed gap, $e_n$ is the local normal-axis basis vector,
and $\mc{K}_{\mu_i}$ and $\mc{K}_{\mu_i}^*$ are the Coulomb friction cone
with coefficient $\mu_i$ and its dual~\cite{lelidec2024contact}.
The complementarity constraint follows Anitescu's cone formulation~\cite{anitescu2006optimization}.
The generalized-inequality KKT
conditions~\cite{boyd2004convex} show that \eqref{eq:cqdc_equation_mat}
is the optimality condition for the following problem:
\begin{equation}
 \min_{\delta q}\ \frac12\|\delta q+\mb{Q}^{-1}b\|_{\mb{Q}}^2
 \quad\mr{s.t.}\quad
 \mb{J}_i\delta q+g_i e_n\in\mc{K}_{\mu_i}^*,\ \forall\,i.
 \label{eq:cqdc_primal}
\end{equation}
Here $\|v\|_{\mb{Q}}^2=v^\top\mb{Q}v$ denotes the squared
$\mb{Q}$-norm.

By neglecting transient dynamics and focusing on transitions between
steady states, the quasi-dynamic approximation supports larger planning
time steps, allowing the planner to look further ahead with fewer
steps~\cite{pang2022global}.  This makes the model particularly useful for
planning quasi-static manipulation tasks dominated by friction and damping.

\subsection{Differentiable Collision Detection}
\label{subsec:diffcollision}

The central idea of sampling-based differentiable collision detection is to
preserve the GJK+EPA forward query~\cite{gilbert1988fast,van2001proximity}
while differentiating its witness points through a smooth surrogate in the
backward pass.  Montaut et al.~\cite{montaut2023differentiable} realize this
idea by formulating the discrete witness-point selection as an argmax and
approximating it with a soft argmax over sampled surface points to make the
witness-point mapping differentiable.  Their direction-based scoring is
restricted to convex geometry and becomes ill-conditioned when the direction
between the witness points vanishes.  DiffCollision~\cite{chen2025robust}
instead writes
the optimality condition in a distance-based form, extending the construction
to concave geometry.  Let $x_b$ be the forward witness point on object $b$ and
$\{p_i^a\}_{i=1}^{N}$ be world-frame surface samples on object $a$.  Its
backward surrogate is
\begin{equation}
 \widetilde x_a=
 \underset{p\in\{p_i^a\}_{i=1}^{N}}{\operatorname{SoftArgmax}_{\tau_a}}
 \bigl(-\|p-x_b\|_2\bigr),
 \label{eq:diffcollision_witness}
\end{equation}
where $\operatorname{SoftArgmax}_{\tau_a}$ denotes the average of the sampled
points weighted by the softmax of their scores with temperature $\tau_a>0$.
DiffCollision also introduces equivalent gradient transport (EGT) to transfer
the gradient of a fixed object to the optimized object when only one pose is
adjustable.

\section{CQDC-Warp}
\label{sec:cqdc_warp}

This section first introduces two modeling improvements of CQDC-Warp over CQDC: a
compliant-contact formulation and a revised inertial model.  We then describe
how gradients are computed through contact geometry using witness-pair
surrogates and through the contact solve via implicit differentiation.  Finally,
we summarize the batched GPU implementation in NVIDIA Warp.

\subsection{Compliant Contact}

We replace the hard-contact model of classical CQDC with a compliant-contact
formulation that retains the friction-cone constraint while penalizing
contact deformation.  The resulting problem is
\begin{equation}
 \begin{aligned}
 \min_{\delta q,\{y_i\}}\quad &
 \frac12\|\delta q+\mb{Q}^{-1}b\|_{\mb{Q}}^2
 +\sum_i\frac12\|y_i-\eta g_i e_n\|_{\mb{R}^{-1}}^2\\
 \mr{s.t.}\quad &\mb{J}_i\delta q+y_i\in\mc{K}_{\mu_i}^*,
 \qquad \forall i,
 \end{aligned}
 \label{eq:cqdc_slack}
\end{equation}
where $\{y_i\}$ are contact-space slack variables, $\eta\in[0,1]$
scales the reference gap, and $\mb{R}=\rho\mb{I}$, $\rho>0$, is the
contact-compliance matrix.

This formulation shares the convex soft-contact structure of
MuJoCo~\cite{todorov2014mujoco,lelidec2024contact} and
SAP~\cite{castro2021sap}, expressed here in displacements rather than
accelerations (MuJoCo) or velocities (SAP).
At the displacement level, the normal response in the stiction branch
acts as a unilateral linear spring: it resists compression without
attraction, with stiffness $1/\rho$ and a zero-force reference gap
$(1-\eta)g_i$.  Thus $\rho$ sets contact softness, while $\eta$ controls
how much existing penetration the spring seeks to recover in one step.
Beyond modeling deformation, finite compliance regularizes the contact problem
and facilitates its numerical solution: as $\rho$ decreases toward the
hard-contact limit, Newton iterations and runtime rise
(Fig.~\ref{fig:timing}(b)).

The shared convex structure also allows the same analytical reduction as MuJoCo:
we can express the optimal $y_i$ as a function of $\delta q$ through cone
projection and substitute it into \eqref{eq:cqdc_slack}.
The resulting unconstrained strictly convex problem is then solved with a
batched GPU Newton method.

\subsection{Revised Inertial Model}

\begin{figure}[t]
 \centering
 \includegraphics[width=\columnwidth]{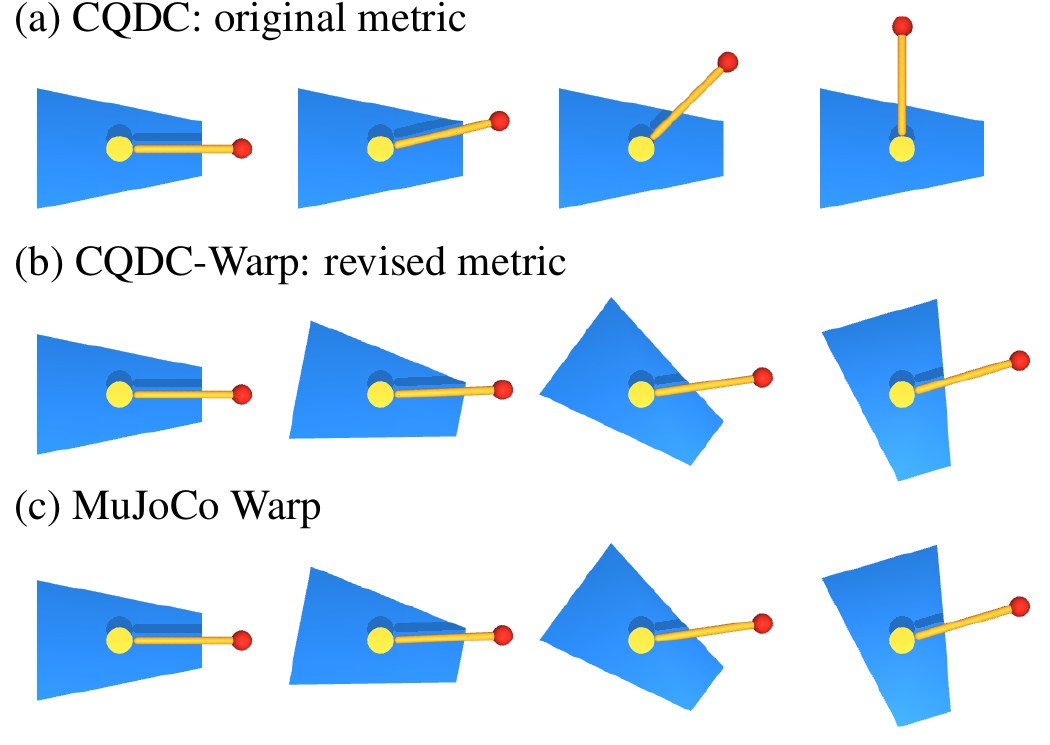}
 \caption{\textbf{Effect of the revised inertial model.} Original
 CQDC metric fails to model joint--base coupling, while our revised metric
 accurately reproduces the trajectory of MuJoCo-Warp.}
 \label{fig:mechanical_metric}
 \vspace{-3mm}
\end{figure}

For floating-base robots, joint motion can induce base motion through
inertial coupling.  The block-diagonal metric of classical CQDC in
\eqref{eq:cqdc_QbJ_definition} omits this coupling, which can contribute to
sim-to-sim and sim-to-real discrepancies, as shown by the toy model in Figure~\ref{fig:mechanical_metric}.
 CQDC-Warp includes inertia at
both actuated and unactuated coordinates and retains their inertial
coupling through the full generalized mass matrix.  To derive the resulting
model, we start from the force balance
\begin{equation}
 \mb{M}(q)a=\mb{K}_{\mr{p}}(u-q-\delta q)+\tau + \mb{J}^\top \lambda,
 \label{eq:cqdc_force_balance}
\end{equation}
where $\mb{M}(q)$ is the full generalized mass matrix, $\mb{K}_{\mr{p}}$ is the
position-control stiffness matrix, nonzero
only at actuated coordinates, $a$ is the generalized acceleration, and
$\tau$ denotes total external forces.  Adopting the step approximation
$a=\delta q/h^2$ and collecting the displacement terms gives the classical
balance form \eqref{eq:cqdc_equation_mat}, with
\begin{equation}
 \mb{Q}(q)=\frac{\mb{M}(q)}{h^2}+\mb{K}_{\mr{p}},\quad
 b(q,u)=\mb{K}_{\mr{p}}(q-u)-\tau.
 \label{eq:cqdc_metric}
\end{equation}

\subsection{Contact-Geometry Surrogate Gradients}
\label{subsec:contact_geometry_surrogates}

DiffCollision independently smooths the two witness points using separate
surface samples and softmax weights.  Although the contact normal and gap can
then be computed differentiably, this two-stage procedure requires propagating
witness-point derivatives to the contact quantities used by the solver.
Instead, we sample witness pairs and use shared
distance-based softmax weights to directly estimate the contact midpoint $c$,
unit normal $n$, and signed gap $g$ together.

Let $(p_i^a,n_i^a)$ denote sampled world-frame surface points and
normals, with superscript $a$ identifying the object.  We pair each forward witness point with samples on the
other object.  For pair $k=(i,j)$, let
\begin{equation}
 d_k=\|p_i^{1}-p_j^{2}\|,\hspace{0.5em}
 n_k=\mathcal N(n_i^{1}-n_j^{2}),\hspace{0.5em}
 c_k=\tfrac12(p_i^{1}+p_j^{2}).
 \label{eq:geometry_pairs}
\end{equation}
where $\mathcal N$ denotes normalization and $n_k$ is oriented to agree with
the forward contact normal.
The weights and three surrogates are
\begin{equation}
 \begin{aligned}
 \alpha_k&=\frac{\exp(-d_k/\tau)}
 {\sum_l\exp(-d_l/\tau)},
 &\widetilde c&=\sum_k\alpha_kc_k, \\
 \widetilde n&=\mathcal N\!\Bigl(\sum_k\alpha_kn_k\Bigr),
 &\widetilde g&=\sum_k\alpha_k(p_j^2-p_i^1)^\top n_k.
 \end{aligned}
 \label{eq:geometry_surrogates}
\end{equation}

Surprisingly, on DiffCollision's witness-point matching
benchmark~\cite{chen2025robust}, our preliminary experiments show that this
formulation matches DiffCollision's performance without EGT, suggesting that
directly approximating contact quantities provides a more natural formulation.

\subsection{Contact-Solve Implicit Gradients}
\label{subsec:contact_solve_gradients}

We differentiate the converged solution $\delta q$ with respect to
$\theta=(\mb{Q},b,\mb{J}_i,g_i)$ by applying the implicit function theorem to the
stationarity equation~\cite{agrawal2019cone,pang2022global}.
Let $r_i=\mb{J}_i\delta q+\eta g_i e_n$ and let $\Pi_i$ denote
Euclidean projection onto $\mc{K}_{\mu_i}^*$.
The stationarity residual of
\eqref{eq:cqdc_slack} is
\begin{equation}
 F(\delta q;\theta)=\mb{Q}\delta q+b+
 \sum_i\mb{J}_i^\top\frac{r_i-\Pi_i(r_i)}{\rho}=0.
 \label{eq:cqdc_stationarity}
\end{equation}
Its displacement Jacobian is
\begin{equation}
 \mb{H}_q\!=\frac{\partial F}{\partial\delta q}\!=\mb{Q}+
 \sum_i\mb{J}_i^\top\mb{C}_i\mb{J}_i,\quad
 \mb{C}_i\in\frac{\mb{I}-\mr{D}\Pi_i(r_i)}{\rho},
 \label{eq:cqdc_hessian}
\end{equation}
where $\mr{D}\Pi_i$ is the cone-projection Jacobian.
For an upstream gradient $\widehat{\delta q}$ written in row vector, 
implicit differentiation solves the following linear system for downstream $\hat\theta$~\cite{agrawal2019cone}:
\begin{equation}
 \widehat\theta=-\widehat{\delta q}\,\mb{H}^{-1}_q\,\frac{\partial F}{\partial\theta}.
 \label{eq:cqdc_adjoint}
\end{equation}

\subsection{Implementation}
\label{subsec:cqdc_warp_implementation}

CQDC-Warp is implemented in Python using NVIDIA Warp~\cite{warp2022} for
GPU-parallel computation and automatic differentiation.  We reuse and adapt
much of MuJoCo-Warp's batched GPU infrastructure, including forward
kinematics, collision detection, and forward
dynamics~\cite{mujocowarp2025}.  Since the MuJoCo-Warp version used in our
implementation does not provide an end-to-end differentiable simulation
pipeline, we augment the backward computation of each component.  For forward
kinematics, we modify selected kernels to enable Warp's automatic adjoint
generation.  Collision detection uses the contact-geometry surrogate gradients
of Sec.~\ref{subsec:contact_geometry_surrogates}, while forward dynamics uses
the contact-solve implicit gradients of
Sec.~\ref{subsec:contact_solve_gradients}.  Both the forward Newton and reverse
adjoint systems are solved in parallel over independent GPU worlds.

\section{CQDC-Warp Model Predictive Control}
\label{sec:cqdc_mpc}

We consider contact-guided local trajectory optimization given a suitable
initialization and intended contacts specified before planning. Contact intent
identifies which robot links should contact the object and enters the MPC
objective through soft contact costs. It does not prescribe a complete
contact-mode schedule: collision detection and the contact solve determine
the realized contact geometry, forces, and frictional response during each
rollout. The optimizer updates command sequences through these dynamics.

\subsection{Batched Receding-Horizon Optimization}

Finite-horizon model predictive control (MPC) solves
\begin{equation}
 \begin{aligned}
 \min_{u_{0:H-1}}\quad &
 \ell_H(x_H)+\sum_{k=0}^{H-1}\ell_k(x_k,u_k,u_{k-1})\\
 \mr{s.t.}\quad &x_0=x_s,\qquad x_{k+1}=f(x_k,u_k).
 \end{aligned}
 \label{eq:generic_mpc}
\end{equation}
Here $x_s$ is the current observed state, $f$ is the dynamics function,
$\ell_k$ and $\ell_H$ are stage and terminal costs, and $u_{-1}$ is the
previously executed command.  MPC executes only the first command $u_0$ and
replans from the next observation~\cite{rawlings2017mpc,mayne2014model}.
We use differentiable CQDC-Warp as the predictive dynamics $f$
in~\eqref{eq:generic_mpc}. At control cycle $s$, we run the following
algorithm.

\begin{algorithm}[H]
 \caption{Batched gradient-based CQDC-Warp MPC}
 \label{alg:cqdc_mpc}
 \begin{algorithmic}[1]
  \Require observed $q_s$, previous solution $U_{s-1}^*$, horizon $H$,
           iterations $N$, step sizes $\{\alpha_j\}$
  \State $U\gets\operatorname{Shift}(U_{s-1}^*)$; initialize
         $(U_{\mr{best}},L_{\mr{best}})$ per world
  \For{$j=0,\ldots,N-1$}
   \State reset all planning worlds to $q_s$
   \For{$k=0,\ldots,H-1$}
    \State $q_{k+1}\gets F(q_k,u_k)$
   \EndFor
   \State $L\gets\mc{L}(U;q_s)$ and $G\gets\nabla_U L$
   \State retain $(U,L)$ where it improves the per-world best
   \State $U\gets U-\alpha_j\mc{B}(G)$
  \EndFor
  \State execute $u_s\gets (U_{\mr{best}})_0$
 \end{algorithmic}
\end{algorithm}

Here $U_{\mr{best}}$ and $L_{\mr{best}}$ store the best sequence and cost
for each GPU world, while $G=\nabla_U\mc{L}$ is the command-sequence
gradient.  The operator $\mc{B}$ normalizes the translation, rotation,
and joint blocks of the gradient separately,
and $\alpha_j$ is the iteration step size.
$\operatorname{Shift}$ drops the executed command and appends a terminal
command to warm-start the next cycle.  After optimization, only the
first command of the best sequence is executed, and the next observation
initializes a new optimization.

\begin{table*}[!t]
\centering
\scriptsize
\setlength{\tabcolsep}{2.2pt}
\resizebox{\textwidth}{!}{%
\begin{tabular}{@{}llcccccccccccccccccc@{}}
\toprule
Method & Metric & Air. & Bin. & Bowl & Bun. & Cam. & Can & Cube & Cup
& Ele. & Foam & Mug & Pig. & Duck & Stick & Tea. & Torus & Bottle & Avg. \\
\midrule
CFree & Succ. (\%)
& 23.2 & 74.6 & 74.6 & 85.7 & 53.2 & 92.1 & 79.1 & 92.5 & 76.0 & 98.6 & \textbf{90.0} & 81.2 & \textbf{90.5} & \textbf{19.1} & 71.5 & 51.6 & \textbf{100.0} & 73.7 \\
& Time (s)
& 22.5 & 21.0 & 21.8 & 20.79 & 23.0 & 21.8 & 21.9 & 22.5 & 22.0 & 19.7 & 20.2 & 22.1 & 25.1 & 22.2 & 20.7 & 20.5 & 16.4 & 21.4 \\
\midrule
IMPACT & Succ. (\%)
& \textbf{24.3} & 65.9 & 75.5 & 67.4 & 62.7 & 94.0 & 45.8 & 97.5 & 71.4 & 88.7 & 74.2 & 63.3 & 55.4 & 6.4 & 83.1 & 44.6 & 88.8 & 65.2 \\
& Time (s)
& 52.2 & 39.9 & 36.2 & 40.3 & 42.8 & 31.3 & 43.0 & 27.0 & 36.5 & 31.8 & 35.5 & 47.9 & 42.7 & 55.5 & 39.3 & 46.6 & 28.7 & 39.8 \\
\midrule
CQDC-Warp & Succ. (\%)
& 20.4 & \textbf{81.7} & \textbf{76.9} & \textbf{91.8} & \textbf{72.2} & \textbf{97.9} & \textbf{83.3} & \textbf{99.3} & \textbf{91.3} & \textbf{98.9} & 84.4 & \textbf{89.5} & 82.2 & 6.0 & \textbf{95.1} & \textbf{86.2} & 99.8 & \textbf{79.8} \\
(Ours) & Time (s)
& \textbf{0.866} & \textbf{0.884} & \textbf{0.880} & \textbf{0.879}
& \textbf{0.894} & \textbf{0.811} & \textbf{0.817} & \textbf{0.901}
& \textbf{0.741} & \textbf{0.732} & \textbf{0.733} & \textbf{0.774}
& \textbf{0.763} & \textbf{0.701} & \textbf{0.696} & \textbf{0.742}
& \textbf{0.685} & \textbf{0.794} \\
\bottomrule
\end{tabular}
}
\caption{\textbf{Allegro in-hand rotation.}
Success and mean trajectory-generation time over $1{,}024$ trials per object.}
\label{tab:allegro_comparison}
\vspace{-3mm}
\end{table*}

\subsection{Trajectory Costs}

Every cost in CQDC-Warp MPC is of the form $P(r)$, where $r$ is the
residual and $P$ applies either L2 or Huber loss to $r$.
We now define the residuals used in our MPC planning framework, grouped into three families.

\subsubsection{Positional Costs}

For entity $e$ at step $k$, translation and rotation residuals are
\begin{equation}
 r^{p}_{k,e}\!=\!\|p_{k,e}-p^*_{k,e}\|_2,\;\;
 r^{R}_{k,e}\!=\!\left\|\operatorname{Log}\!\left(
 (\mb{R}^*_{k,e})^\top\mb{R}_{k,e}\right)\right\|_2,
 \label{eq:pose_errors}
\end{equation}
where a star denotes the goal and $\operatorname{Log}$ maps the relative
rotation to its axis--angle vector.  These residuals directly track the
desired pose.  To compensate for insufficient positional gradients near goal,
we also penalize motion away from goal using the linear and angular
goal-error rates
\begin{equation}
 \begin{aligned}
 r^{\dot p}_{k,e}\!=\!\!\left[\!\frac{e^{p}_{k,e}\!\!
 -\!\sg(e^{p}_{k-1,e})}{h}\!\right]_+
 r^{\dot R}_{k,e}\!=\!\!\left[\!\frac{e^{R}_{k,e}\!\!
 -\!\sg(e^{R}_{k-1,e})}{h}\!\right]_+
 \end{aligned}
 \label{eq:goal_error_rates}
\end{equation}
Here $\sg(\cdot)$ denotes stop-gradient.  Detaching the
previous error preserves the new-endpoint gradient and prevents the rate
from being reduced by moving the previous endpoint.

\subsubsection{Regularization Costs}

Linear and angular velocity residuals activate above configured limits,
while command velocity and acceleration regularize command magnitude and
temporal variation:
\begin{equation}
 \begin{aligned}
 r^v_{k,e}&=[\|v_{k,e}\|_2-\bar v_e]_+,&
 r^\omega_{k,e}&=[\|\omega_{k,e}\|_2-\bar\omega_e]_+,\\
 r^{\dot u}_k&=u_k/h,&
r^{\ddot u}_k&=(u_k-u_{k-1})/h^2.
 \end{aligned}
 \label{eq:regularization_residuals}
\end{equation}
Here $v_{k,e}$ and $\omega_{k,e}$ are the entity's linear and angular
velocities; $\bar v_e$ and $\bar\omega_e$ are their respective limits.
For joint coordinate $q_{k,j}$ with limits
$[\underline q_j,\overline q_j]$ and safety margin $m_j$, define the
residuals
\begin{equation}
 r^{\mr{jl}}_{k,j,-}\!\!=\![\underline q_j+m_j-q_{k,j}]_+,\quad
 r^{\mr{jl}}_{k,j,+}\!\!=\![q_{k,j}-\overline q_j+m_j]_+.
 \label{eq:joint_limit_residual}
\end{equation}

\subsubsection{Contact Costs}

For each task, we specify the robot links intended to make contact before
planning.  For each designated contact $c$, the contact distance residual
measures the deviation of the signed gap $g_{k,c}$ from its target $g^*$.
Optionally, the contact position residual anchors the witness
point $c_{k,c}$ to a fixed world-frame reference $c_c^*$ along selected
Cartesian axes $\mc{A}\subseteq\{x,y,z\}$.  The residuals are
\begin{equation}
 r^{\mr{cd}}_{k,c}=g_{k,c}-g^*,\quad
 r^{\mr{cp}}_{k,c}=\|(c_{k,c}-c_c^*)_{\mc{A}}\|_2.
 \label{eq:contact_residuals}
\end{equation}
The distance cost encourages the intended contact, while the optional
position cost biases its location along selected axes. Both guide contact
realization without fixing the witness point or frictional mode in the
dynamics.

\section{Experiment}
\label{sec:experiments}

\begin{figure*}[t]
 \centering
 \includegraphics[width=\textwidth]{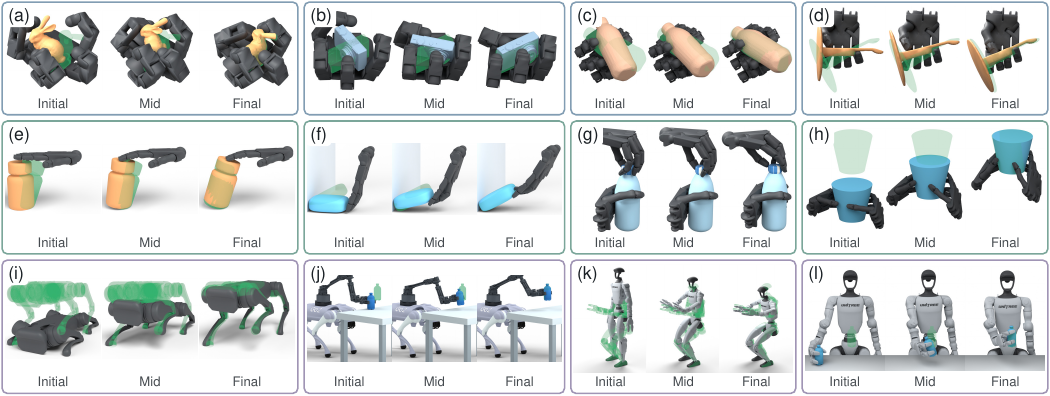}
 \caption{\textbf{Representative contact-rich trajectories, grouped by task:}
 (a)--(d) in-hand rotation with Allegro, LEAP, and Shadow hands;
 (e)--(f) environmental contact exploitation using table, wall, and floor
 contacts; (g)--(h) bimanual bottle opening and object lifting; and
 (i)--(l) whole-body control with A1, GO2-D1, and G1.
 Light-green shapes indicate goal poses.}
 \label{fig:unified_experiments}
 \vspace{-3mm}
\end{figure*}

\textbf{Common Setup:}
All CQDC-Warp planning uses one NVIDIA RTX 4090 GPU, with shared parameters
within each task family. Each control cycle uses $20$ gradient iterations
and horizon $H=5$. Initial step sizes for the translation, rotation, and
joint blocks are $10^{-4}$, $10^{-3}$, and $10^{-2}$, respectively, with
cosine decay to $10\%$. Planning and MuJoCo evaluation time steps are
$0.01$ and $0.002\,$s. Each rollout runs for $500$ control steps without
early stopping. Unless specified otherwise, success requires all tracked
position and geodesic rotation errors to remain below $2\,$cm and $5^\circ$
for $20$ consecutive evaluation steps.

\subsection{In-Hand Rotation}
\label{subsec:in_hand_rotation}
\textbf{Allegro Hand.}
We use the 17-object CFree benchmark~\cite{jin2024complementarity}, also
used by IMPACT~\cite{li2026impact}, with $1{,}024$ trials per object sampled
from its initial-position and target-pose distribution. For this benchmark, 
the success threshold for rotation is $15^\circ$. CQDC-Warp plans all
$1{,}024$ worlds in parallel,
while CFree and IMPACT each run sequentially on one CPU core.

Over $17{,}408$ trials per method (Table~\ref{tab:allegro_comparison}),
CQDC-Warp achieves $79.8\%$ mean success, versus $73.7\%$ for CFree and
$65.2\%$ for IMPACT. Mean trajectory-generation time is $0.794\,$s,
versus $21.422\,$s and $39.844\,$s, corresponding to $27.0\times$ and
$50.2\times$ speedups under these execution settings.

\textbf{LEAP and Shadow Hands.}
Using the Allegro parameters, we evaluate eight object-rotation tasks each
on the 16-DoF LEAP and 22-DoF Shadow hands, plus a
$90^\circ$--$180^\circ$ Shadow box flip. Initial hand--object configurations
come from Dexonomy and BODex~\cite{chen2025dexonomy,chen2025bodex}, following
dataset-based initialization~\cite{wang2022dexgraspnet}. Object-rotation targets
perturb the initial orientation about a uniformly sampled axis by an angle
uniformly sampled from $15^\circ$--$30^\circ$. Success rates are $79.88\%$
for LEAP, $65.75\%$ for Shadow, and $54\%$ for the box flip
(Fig.~\ref{fig:teaser}), supporting transfer across hand embodiments.

\subsection{Additional Robots and Tasks}
\label{subsec:additional_robots_and_tasks}
\textbf{Environmental Contact Exploitation.}
The Shadow Hand tilts a bottle resting on a table by applying frictional
forces from above, or raises the near end of a rounded rectangular object
resting against a wall on the floor. Both tasks use environmental contacts
as passive support and require maintaining hand--object contact without an
enveloping grasp. With identical cost weights and MPC parameters, CQDC-Warp
succeeds in $938/1024$ bottle trials and $902/1024$ wall-supported trials.

\textbf{Bimanual Manipulation.}
Two independently actuated Shadow hands coordinate to open a bottle or
lift an object. For opening, the left hand holds the bottle body at a fixed
position while the right rotates its articulated cap by $90^\circ$,
achieving $1024/1024$ successes. For lifting, the hands grasp opposite sides
of a solid truncated cone and raise it by $0.20\,$m while preserving its
orientation, achieving $1007/1024$ successes.

\textbf{Whole-Body Control.}
We evaluate four tasks using sparse pose-tracking objectives:
\textbf{A1} rises from prone to a three-legged stance, tracking only the
trunk and raised front-left leg;
\textbf{GO2 with a D1 arm} lifts a bottle from a table while maintaining
its base pose, tracking only the bottle and base;
\textbf{G1} reaches and maintains a half-sitting posture from standing,
tracking the torso, pelvis, wrists, and ankles, or lifts a bottle from a
table, tracking only the bottle and torso.

\begin{figure}[t]
 \centering
 \includegraphics[width=0.9\columnwidth]{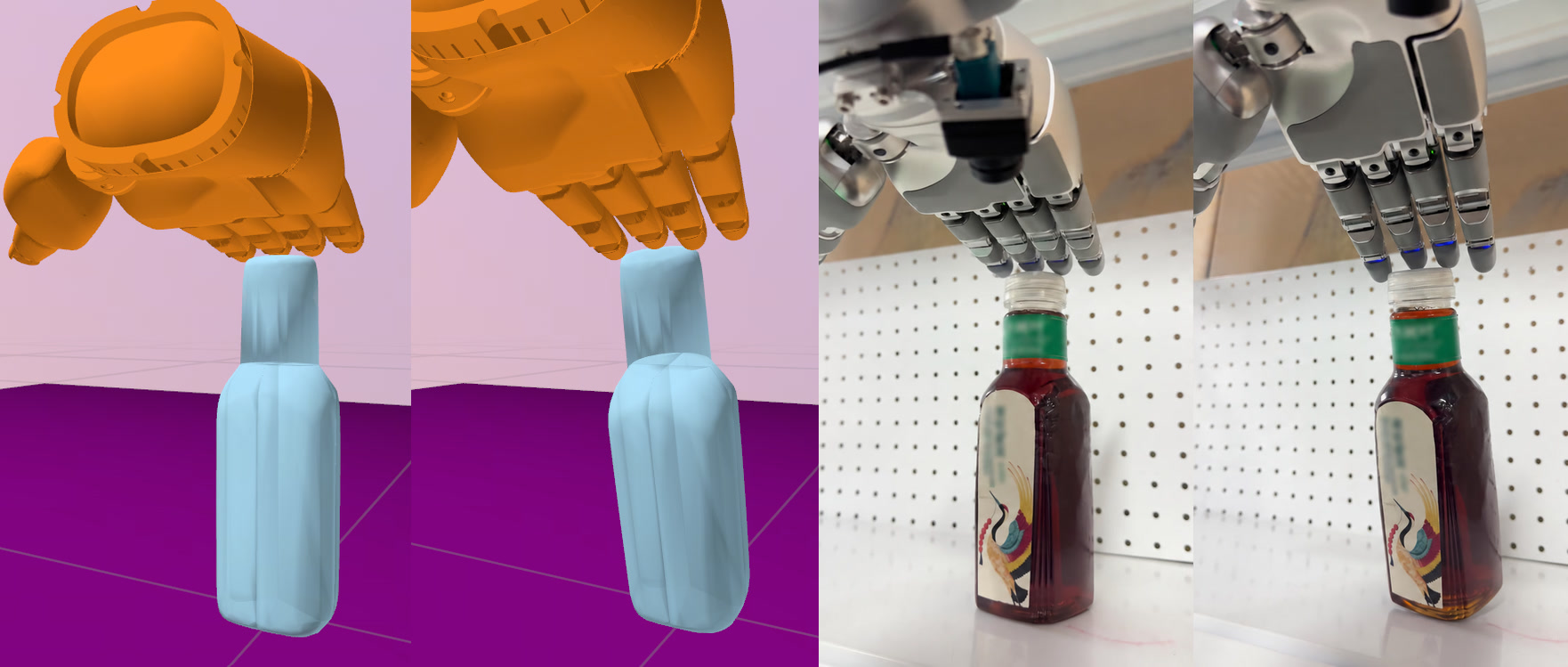}
 \caption{\textbf{Real-world trajectory replay.}
 We replay a CQDC-Warp trajectory to tilt a
 bottle with table contact.}
 \label{fig:sharpa_table}
 \vspace{-3mm}
\end{figure}

\textbf{Real-World Replay.}
We transfer the bottle--table task to a physical Sharpa hand using its model
in planning. We obtain the manipulated-object model by scanning the target
bottle and reconstructing its digital twin. The robot successfully replays
the CQDC-Warp trajectory at
$10\,$Hz over five times its simulated duration (Fig.~\ref{fig:sharpa_table}),
supporting the trajectory's physical plausibility and its potential use for
robot learning.

\subsection{Time Analysis}
\label{subsec:time_analysis}

We measure throughput in the Allegro bunny MPC loop over $100$ control
cycles at batch sizes $\{128,256,512,1024,2048\}$, using the common setup
of Sec.~\ref{sec:experiments}. Each world executes $10{,}000$ forward and
backward steps; throughput counts world-steps per second.
The forward pass includes kinematics, dynamics, collision detection,
constraint assembly, contact solving, and integration. The backward pass
includes local-point sampling for geometry-aware gradients and differentiation
through the cost and simulator. MuJoCo-Warp~\cite{mujocowarp2025} provides
a forward reference on the same scene.

\begin{figure}[t]
 \centering
 \includegraphics[width=\columnwidth]{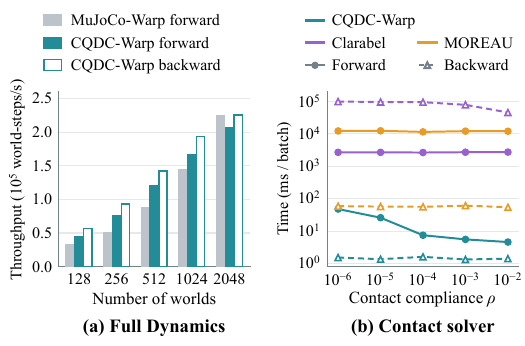}
 \caption{\textbf{Computational efficiency.}
 }
 \label{fig:timing}
 \vspace{-3mm}
\end{figure}

Figure~\ref{fig:timing}(a) shows comparable forward throughput for CQDC-Warp
and MuJoCo-Warp across all batch sizes. At $2048$ worlds, they reach
$2.06\times10^5$ and $2.24\times10^5$ world-steps/s, respectively, while
CQDC-Warp backward reaches $2.25\times10^5$ world-steps/s. 

To isolate solver performance, we solve ten problems extracted from saved
runs, each with batch size $1{,}024$, with $\mb{R}=\rho\mb{I}$ and
$\rho\in\{10^{-6},10^{-5},10^{-4},10^{-3},10^{-2}\}$. We compare
CQDC-Warp, GPU MOREAU~\cite{moreau2026}, and CPU
Clarabel~\cite{goulart2026clarabel}, using
CVXPYLayers~\cite{agrawal2019differentiable} for Clarabel's backward pass.
As $\rho$ increases (i.e., the contact becomes softer), CQDC-Warp forward time falls from $47.97$ to
$4.60\,$ms and mean Newton iterations from $14.83$ to $3.41$
(Fig.~\ref{fig:timing}(b)), indicating that lower compliance
increases solver difficulty. At $\rho=10^{-4}$, combined forward and
backward time is $9.07\,$ms, versus $97.52\,$s for Clarabel and $11.54\,$s
for MOREAU: speedups of $1.08\times10^4$ and $1.27\times10^3$, respectively.

\subsection{Ablation Study}
\label{subsec:ablation}
We use the 17-object Allegro benchmark
(Sec.~\ref{subsec:in_hand_rotation}) here.

\textbf{Geometry-Aware Gradient.} We disable our contact-geometry derivative and fix witness points in each object's local frame. This reduces
success from $79.81\%$ to $77.37\%$, a $2.44$-percentage-point drop,
indicating that contact-geometry derivative helps contact position adjustment.

\textbf{Horizon.}
We vary $H\in\{1,3,5,7,9\}$
(Table~\ref{tab:horizon_ablation}). The default $H=5$ achieves the highest
success ($79.77\%$); longer horizons increase generation time without
improvement. Success falls to $69.39\%$ at $H=3$ and zero at $H=1$,
suggesting that lookahead is necessary to anticipate object motion.

\begin{table}[t]
\centering
\footnotesize
\setlength{\tabcolsep}{3pt}
\begin{tabular}{@{}lccccc@{}}
\toprule
$H$ & 1 & 3 & 5 & 7 & 9 \\
\midrule
Success (\%) & 0.00 & 69.39 & \textbf{79.77} & 79.20 & 78.34 \\
Time (s) & \textbf{0.76} & 1.09 & 1.31 & 1.41 & 1.74 \\
\bottomrule
\end{tabular}
\caption{\textbf{Planning-horizon ablation.}
Success and mean trajectory-generation time with $512$ worlds per object.}
\label{tab:horizon_ablation}
\vspace{-3mm}
\end{table}

\section{Conclusion and Limitations}
\label{sec:conclusions}

We presented CQDC-Warp, a GPU-accelerated differentiable quasi-dynamic
simulator for contact-rich manipulation. It supports the GPU parallelization of 
our contact-guided MPC formulation with intended contacts to guide planning while leaving their
realization to the simulated dynamics. Experiments across versatile robotic tasks
demonstrate the framework's effectiveness and efficiency for generating
diverse contact-rich trajectories.

Our work has three main limitations. First, the planner relies on suitable
initializations and intended contacts specified before planning. Further
evaluation is needed on long-horizon tasks with frequent contact transitions.
Second, although parameters are shared within each task family, objective
weights still require tuning across different tasks and robot embodiments.
Finally, the use of CQDC-Warp for large-scale manipulation data generation
and its impact on downstream policy learning remain to be evaluated.

\bibliographystyle{IEEEtran}
\bibliography{IEEEabrv}

\end{document}